\documentclass{article}

\usepackage{arxiv}

\usepackage[utf8]{inputenc} % allow utf-8 input
\usepackage[T1]{fontenc}    % use 8-bit T1 fonts
\usepackage{hyperref}       % hyperlinks
\usepackage{url}            % simple URL typesetting
\usepackage{booktabs}       % professional-quality tables
\usepackage{amsfonts}       % blackboard math symbols
\usepackage{nicefrac}       % compact symbols for 1/2, etc.
\usepackage{microtype}      % microtypography
\usepackage{lipsum}		% Can be removed after putting your text content
\usepackage{graphicx}
\usepackage{natbib}
\usepackage{mathtools}
\usepackage{doi}

\title{n-stage Latent Dirichlet Allocation: A Novel Approach for LDA}

\author{ \href{https://orcid.org/0000-0002-7025-2815}{\includegraphics[scale=0.06]{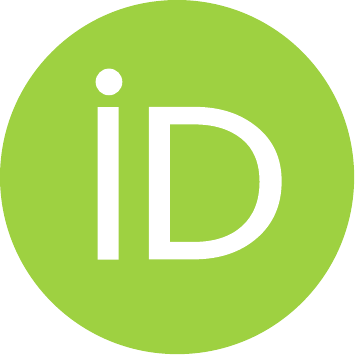}\hspace{1mm}Zekeriya Anil Guven}\thanks{Ege University, Faculty of Engineering, Computer Engineering Department, Izmir, anilguven1055@gmail.com. This study is extension version of "Comparison of Topic Modeling Methods for Type Detection of Turkish News" (\url{http://dx.doi.org/10.1109/UBMK.2019.8907050}). Please citation this IEEE paper.} \\
	Department of Computer Engineering\\
	Ege University\\
	Izmir, Turkey \\
	\texttt{anilguven1055@gmail.com} \\
	\And
	\href{https://orcid.org/0000-0002-4052-0049}{\includegraphics[scale=0.06]{orcid.pdf}\hspace{1mm}Banu Diri} \\
	Department of Computer Engineering\\
	Yildiz Technical University\\
	Istanbul, Turkey \\
	\texttt{diri@yildiz.edu.tr} \\
	\And
	\href{https://orcid.org/0000-0002-4711-7287}{\includegraphics[scale=0.06]{orcid.pdf}\hspace{1mm}Tolgahan Cakaloglu} \\
	Walmart Global Tech\\
	Dallas, USA \\
	\texttt{jackalhan@gmail.com} \\
}

\renewcommand{\shorttitle}{\textit{arXiv} Template}

\hypersetup{
pdftitle={A template for the arxiv style},
pdfsubject={q-bio.NC, q-bio.QM},
pdfauthor={David S.~Hippocampus, Elias D.~Striatum},
pdfkeywords={First keyword, Second keyword, More},
}

\begin{document}
\maketitle

\begin{abstract}
	Nowadays, data analysis has become a problem as the amount of data is constantly increasing. In order to overcome this problem in textual data, many models and methods are used in natural language processing. The topic modeling field is one of these methods. Topic modeling allows determining the semantic structure of a text document. Latent Dirichlet Allocation (LDA) is the most common method among topic modeling methods. In this article, the proposed n-stage LDA method, which can enable the LDA method to be used more effectively, is explained in detail. The positive effect of the method has been demonstrated by the applied English and Turkish studies. Since the method focuses on reducing the word count in the dictionary, it can be used language-independently. You can access the open-source code of the method and the example: \url{https://github.com/anil1055/n-stage\_LDA} 
\end{abstract}

% keywords can be removed

\section{Introduction}
Due to the increase of electronic document archives, new techniques or tools must be used to automatically organizing, searching, indexing and scanning large collections to manage documents. However, as manual analysis of the data in the document becomes more and more difficult, text mining methods have been developed to analyze the data automatically. Text mining is used to identify the core of topic discovery, customer relationship management, and target advertising in the information systems discipline. It is also used to detect textual objects from images. Considering the need to analyze and understand text data, it is important to analyze and apply text mining \citep{Lee2010}. One of the important methods used in text mining is topic modeling (TM) algorithms. TM is a machine learning method for natural language processing that allows determining the semantic structure of a text document \citep{Blei2012}. The purpose of TM is to explore how to combine documents that share a word usage or similar models. Therefore, topic models can be studied with documents. These documents can be thought of as a mix of topics. The topic is the probability distribution on a word. In other words, TM is a generative model for documents \citep{Alghamdi2015}. From a broad perspective, TM methods are applied to natural language processing, text mining, social media analysis, information retrieval. For example, TM based on social media analysis makes it easier to understand reactions and conversations between users in applications such as Twitter, Facebook. Topic models are a prominent method for displaying discrete data and offer an efficient approach to finding hidden structures in gigantic information \citep{Jelodar2019}.

One of the most common algorithms for TM is the Latent Dirichlet Allocation (LDA) algorithm. The basic idea of the LDA method is that each document exhibits a mix of hidden topics, where each topic is characterized by a distribution by words, and the relative importance of topics varies from document to document \citep{BASTANI2019}. The LDA algorithm considers all word tokens to be of equal importance, it is a probabilistic model. It is useful for TM, classification, collaborative filtering and information extraction \citep{Martin2015}. 

In this article, the LDA method is explained in detail. Then a novel approach, namely n-stage LDA (n-LDA), is proposed for more effective usage of LDA. The purpose of n-LDA method is to reduce the size of LDA's dictionary and increase the weight of related words to make topic labelling easier. The positive effects of this new approach are explained and shown in the next sections.

The content of this article is as follows. In the second section of this paper, the related works for LDA are explained. The LDA and proposed n-stage LDA methods are mentioned in the third section. In the fourth section, applied works for n-stage LDA are indicated and the positive effect of the n-stage LDA methods are shown. In the last section, the conclusions of this study are explained.

\section{Related Works}

Topic models like LDA are applied in various fields including medical sciences, social medias, software engineering, geography, political science, etc.

\cite{tran2019} analyzed the evolution of global trends, patterns and interdisciplinary landscapes in artificial intelligence and cancer research. They used the LDA to classify papers by relevant topics. \cite{Xue2020} aimed to understand Twitter users' discourse and psychological reactions to COVID-19. They used machine learning techniques and LDA to analyze approximately 1.9 million English Tweets about the coronavirus. Sentiment analysis showed fear towards the unknown nature of the coronavirus. \cite{vikram2019} utilized both visual and textual methods to improve image retrieval in medical data. They proposed a LDA-based technique to encode visual features and showed that these features effectively model medical images.

\cite{Zhou2021} proposed a guided LDA workflow for investigating temporary hidden topics in tweets during a recent catastrophic event, Hurricane Laura 2020. With the integration of preliminary information, visualization of LDA topics and validation from official reports, their guided approach revealed that most tweets contained several hidden topics during the 10-day period of Hurricane Laura. \cite{jia2019toward} reviewed online ratings and reviews of nearly 54,000 pairs of exercisers from 100 fitness clubs in Shanghai, China. They used LDA-based text mining to identify 17 main topics that the exercisers had written on. \cite{Hidayatullah2017} aimed to establish the topic model of traffic information in Indonesian Twitter messages. The data used were taken from the official Twitter account of the Traffic Management Center in Java and used the LDA method to generate the topic model.

\cite{de2021lda} propose a new approach to this task, using the LDA method to identify the main topics underlying each political speech and distribute relevant terms among different topic-based sub-profiles. \cite{Bertalan2019} proposed an TM framework using LDA and other TM methods to identify the main topics of discussion on political websites.

\cite{Putri2017} used the LDA method to assess the trend from tourist review to specific topics that can be classified correctly, positive and negative emotions, and proposed sentiment analysis with a probabilistic topic model. \cite{Guven2019b} used TM methods to determine which headline type the Turkish news headlines belong to. They analyzed LDA, n-stage LDA and other TM methods and compared their results.

\section{Methods}
\subsection{Latent Dirichlet Allocation}

LDA was first developed by \cite{Blei2003} as a generative probabilistic modeling approach to reveal hidden semantic structures in a collection of textual documents. LDA is an algorithm for text mining based on a statistical (Bayesian) topic model and is very widely used. Thus, the LDA creates a document with the determined topics. In the LDA, each document is modeled as a mix of topics. Each topic represents a different probability distribution that defines the probability of each word appearing on a particular topic \citep{Alghamdi2015}. With the LDA, each document is represented as a multinomial distribution of topics, where topics can be viewed as higher-level concepts similar to clusters. This method is based on the assumption that each document in a collection is composed of several hidden topics in which each topic is presented \citep{Pavlinek2017}. 

LDA is an effective, unsupervised learning method used for topic detection. The method uses a collection of multiple unlabeled documents \citep{Bolelli2009}. The topic and word-weight distributions of the document modeled by the LDA show the best topics for the document \citep{Momtazi2018}.

With LDA, a random topic is assigned to all words in the document. Using this information, certain statistics are extracted. Local statistics indicate how many words are assigned to topics in each document. The global statistics show how many times each word in the whole document is assigned to the topics. With statistical information, a new topic is assigned to each word for each document \citep{Guven2018}. 

The model generation process for each document in the text archive is shown as follows \citep{Chen2019}.

\begin{figure}[h]
	\centering
	\includegraphics[width=8.0cm]{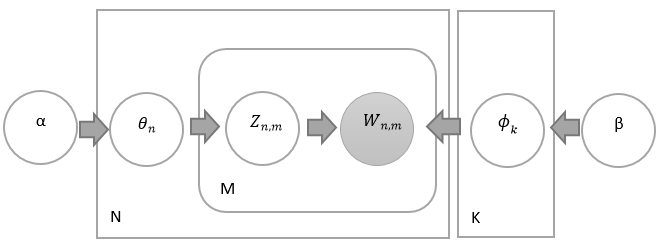}
	\caption{Structure of LDA \citep{Blei2003}}
	\label{fig:fig1}
\end{figure}

\begin{itemize}
	\item $\theta_{n}$ for the nth of documents d in the entire collection of N documents - Dirichlet($\alpha$) selection
	\item For each $W_{n,m}$ word in document d:
	\subitem $Z_{n,m}$ - Multinomial($\theta_{n}$) topic assignment is selected
	\subitem $\phi_Z$ $_{n,m}$ - Dirichlet($\beta$) related topic distribution is found
	\subitem The word $W_{n,m}$ is sampled - Multinomial($\phi_Z$ $_{n,m}$).
\end{itemize}

The above steps detail the productive process for all N documents in corpus D and for each document. All operations are repeated N times. The process is graphically illustrated in Figure \ref{fig:fig1} ($\alpha$ and $\beta$ are two Dirichlet-priority hyperparameters) \citep{Chen2019}.

\subsection{n-stage Latent Dirichlet Allocation}

The n-stage LDA method has been proposed to increase the success of the system modeled with the LDA algorithm. The n-LDA method is aimed to delete the words that negatively affect the accuracy of the system from the dictionary. As a result, it is predicted that the weight values of the remaining words will increase and the class labels of the topics can be determined more easily. The reason why we call the proposed system as n stages is that it is dynamic according to the size of the dataset used in the system. If n is 1, it is classical LDA. The value of n is incremented linearly starting from 2.

\begin{figure}[h]
	\centering
	\includegraphics[width=12.0cm]{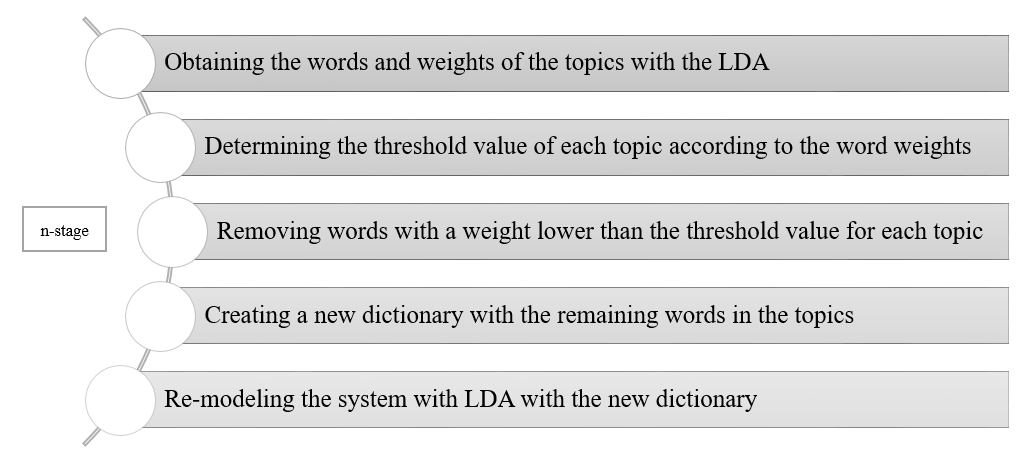}
	\caption{The stages of n-LDA method.}
	\label{fig:fig2}
\end{figure}

The steps of the method are shown in Figure \ref{fig:fig2}. In order to decrease the word count in the dictionary, the threshold value for each topic is calculated. The threshold value is obtained by dividing the sum of the weights of all the words by the word count in the relevant topic. Words with less weight than the specified threshold value are deleted from the topics. Then, a new dictionary is created with the remaining words for the n-LDA model. Finally, the system is re-modeled using the LDA algorithm with the new dictionary. These steps can be repeated n times \citep{Guven2019}.

As the number of steps in the method increases, the number of words in the dictionary decreases. Therefore, the weight values of the remaining words in the dictionary change positively. The positive effect of the proposed method has been demonstrated in Section \ref{sectionApplied}.

\begin{figure}[h]
	\centering
	\includegraphics[width=10.5cm]{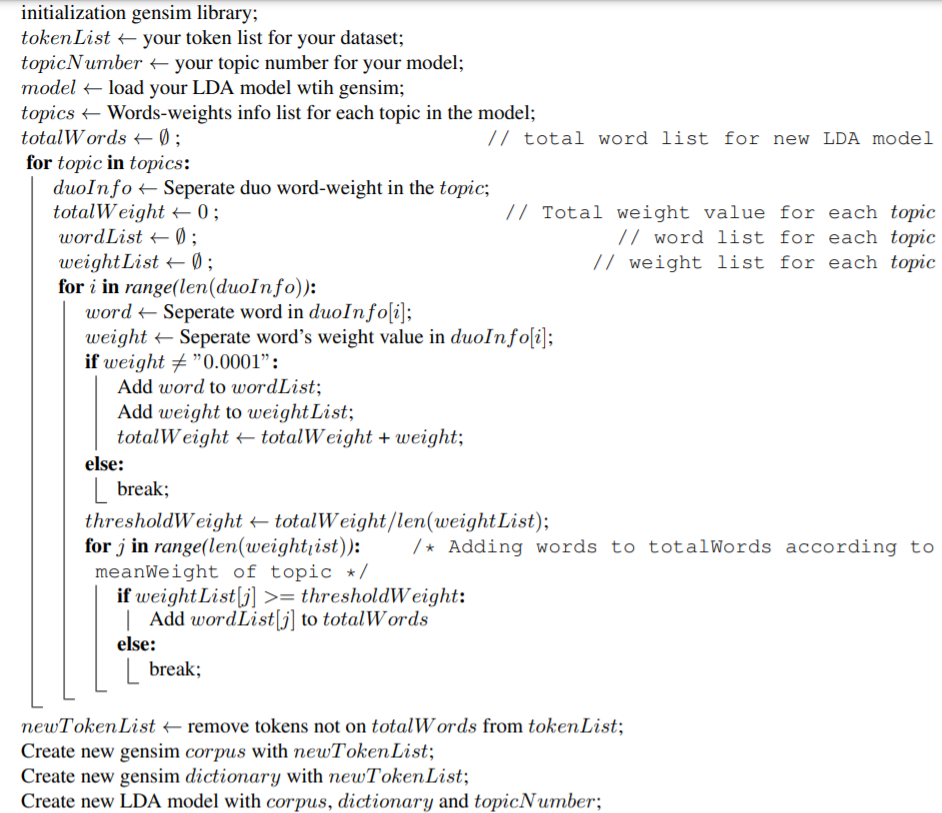}
	\caption{The pseudocode of n-LDA method.}
	\label{fig:fig3}
\end{figure}

The pseudocode of the n-LDA method is given in Figure 3. You can also access the method's code and example via github site \footnote{https://github.com/anil1055/n-stage\_LDA}.

\section{Applied Works for n-stage LDA}
\label{sectionApplied}

The n-LDA method has been used in many studies and its positive effect has been shown with results. Sentiment analysis, sentiment detection, classification of news headlines can be given as examples to these studies. These studies are listed below:

\begin{itemize}
	\item \textbf{Classification of Turkish Tweet emotions by n-stage Latent Dirichlet Allocation}: The LDA was used to determine which emotions the tweets in this study. Dataset consists of anger, fear, happiness, sadness and surprise, 5 emotion labels and 4000 tweets. Zemberek, Snowball and the first 5 letter root extraction methods are used to create the model. The generated models were tested with the proposed n-stage LDA method and compared with the LDA. For the 5 classes of normal LDA method, the highest 60.4\% success was achieved; 70.5\% for 2-stage LDA and 76.4\% for 3-stage LDA \citep{Guven2018a}.
	\item \textbf{Classification of New Titles by Two Stage Latent Dirichlet Allocation}: In this study, it is aimed to be able to determine which headlines type belongs to English and Turkish news headlines collected from news sites. For this, the proposed two-stage LDA method, which is based on the LDA algorithm, was used in the model. The LDA and 2-LDA method was analyzed and compared with each other. Then, by creating a file with an arff extension from the word weights of the topics, the success of the machine learning methods in Weka was measured. Random Forest was the best successful method for both datasets \citep{Guven2018}.
	\item \textbf{Comparison of Topic Modeling Methods for Type Detection of Turkish News}: This paper, it is aimed to determine which types of news titles belong. The dataset consists of 4200 Turkish new titles belonging to 7 class labels. In order to determine the headline types, classical LDA, Latent Semantic Analysis (LSA) and Non-Negative Matrix Factorization (NMF) algorithms were used in TM. In addition, the proposed n-stage LDA method was also analyzed for this task. The accuracy of all methods was measured and compared. NMF was the most successful method for three classes, while for five and seven classes LSA was the most successful method. On the other hand, n-LDA has achieved more successful accuracy than LDA \citep{Guven2019b}.
	\item \textbf{Emotion Detection with n-stage Latent Dirichlet Allocation for Turkish Tweets}: In this study, LDA and n-stage LDA methods were used to determine which emotions Turkish tweets have. The dataset consists of 5 labels: anger, fear, happiness, sadness and surprise. The accuracy of the models created using rooting methods was analyzed for both methods and the results were compared. With the application of n-LDA, the number of words decreases and the weight of some words increases. Thus, a more successful analysis is applied. When the results were examined, the n-LDA method was more successful than LDA \citep{Guven2019a}.
	\item \textbf{Comparison Method for Emotion Detection of Twitter Users}: In this study, the LDA, n-stage LDA and Non-Negative Matrix Factorization (NMF) methods in TM were used to determine the emotions of Turkish tweets. The dataset consists of 5 emotions; anger, fear, happiness, sadness and surprise. The accuracy of all TM methods used in the study was analyzed and compared with each other. As a result, while the most successful method was NMF (89.6\%),  the most unsuccessful method was the classical LDA (65.8\%, 3-LDA: 81.5\%). Then, the F1 measure of the machine learning algorithms was analyzed by creating a file according to Weka with word weights and class labels of the topics. When the results were examined, the most successful method was n-stage LDA (97.8\%) while the most successful algorithm was Random Forest \citep{Guven2019} (in English arXiv pre-print: \url{https://arxiv.org/abs/2110.00418}).
	\item \textbf{Comparison of n-stage Latent Dirichlet Allocation versus other topic modeling methods for Emotion Analysis}: In this article, LDA, Latent Semantic Analysis (LSA) and Probabilistic-Latent Semantic Analysis (P-LSA) were used to determine the emotions of individuals from Turkish tweets. In addition, the success of the proposed n-stage LDA, which is based on the LDA, algorithm in the emotion analysis was analyzed and compared with the existing methods. The dataset consists of 4000 tweets of 5 different emotions, including anger, fear, happiness, sadness and surprise. All TM methods were modeled for 3 and 5 class datasets and their accuracy values and running times were analyzed. It has been observed that the n-stage LDA (76.375\%) method achieves success in terms of running time and performance according to LDA (60.375\%) and P-LSA (63.125\%). But the most successful and fastest modeled method was LSA (87\%) \citep{Guven2020}.
\end{itemize}

\section{Conclusion and Future Works}

In this article, the n-stage LDA method, which enables the use of the LDA method as n-stage, is explained in detail. With the method, by reducing the word count used by LDA in the dictionary for every stage, successful TM can be done with more accurate topic weighting. The effect of the method in the English and Turkish language is indicated the previous section. Since the method focuses on reducing the word count in the dictionary, it can be used other languages as language-independently.

The use of the n-LDA method in other areas of natural language processing is also considered in future works. Spam detection, sarcasm detection, etc. examples of these fields.

\section{Acknowledgment}

I would like to express my gratitude to Prof. Dr. Banu Diri, who has made a great contribution to me and supported me throughout my life, and also Tolgahan Cakaloglu, who has always provided his help and support in all matters.

\bibliographystyle{unsrtnat}
\bibliography{references}  
\end{document}